\documentclass{article}

\usepackage[preprint]{neurips_2026}

\usepackage[utf8]{inputenc} 
\usepackage[T1]{fontenc}    
\usepackage{hyperref}       
\usepackage{url}            
\usepackage{booktabs}       
\usepackage{amsfonts}       
\usepackage{nicefrac}       
\usepackage{microtype}      
\usepackage{xcolor}         
\usepackage{amsmath}        
\usepackage{enumitem}       
\usepackage{multirow}       
\usepackage{algorithm}
\usepackage{algpseudocode}

\usepackage{comment}

\usepackage{graphicx} 

\usepackage{colortbl}

\usepackage{xcolor}
\definecolor{intraBlue}{RGB}{70,130,180}     
\definecolor{interGreen}{RGB}{76,153,125}

\usepackage{subcaption}
\usepackage{wrapfig}

\hypersetup{
	colorlinks=true,
	citecolor=cyan,
}

\usepackage{pifont}
\usepackage{enumitem}
\let\oldding\ding
\renewcommand{\ding}[2][1]{\scalebox{#1}{\oldding{#2}}}

\definecolor{vbenchblue}{RGB}{235,245,255}
\definecolor{efforange}{RGB}{255,245,235}
\definecolor{paradigmgray}{RGB}{235,235,235}
\definecolor{oursgreen}{RGB}{228,242,228}
\newcommand{\gc}{\cellcolor{oursgreen}}

\title{Attend Locally, Remember Linearly: \\ Linear Attention as Cross-Frame Memory for Autoregressive Video Diffusion}

\author{%
  Kunyang~Li \quad Mubarak~Shah \quad Yuzhang~Shang \\
  Institute of Artificial Intelligence, University of Central Florida
}

\begin{document}
\maketitle

\vspace{-2.5em}
\begin{center}
  \href{https://lky-ang.github.io/ARL2/}{\textcolor{magenta}{\textbf{Project Page}}} \qquad
  \href{https://github.com/lky-ang/ARL2}{\textcolor{magenta}{\textbf{Code}}}
\end{center}

\begin{abstract}
Autoregressive (AR) video diffusion is a powerful paradigm for streaming and interactive video generation. However, its reliance on softmax self-attention leads to quadratic compute complexity in sequence length and increasing memory usage due to key-value caching, which fundamentally limits its scalability to long video horizons.
Existing remedies (e.g., sparse attention and KV-cache compression) reduce per-step cost but still rely on a linearly growing cache or irreversibly discard past context, and thus fail to simultaneously address linear memory growth and streaming context management.
To address this scalability bottleneck, we propose ARL$^2$ \emph{(\textbf{A}ttend \textbf{L}ocally, \textbf{R}emember \textbf{L}inearly)}, a hybrid attention module that replaces quadratic cross-frame attention with a fixed-size recurrent state. Specifically, we decompose self-attention into two branches: an \emph{intra-frame} softmax branch for spatial detail and local dependencies, and an \emph{inter-frame} gated recurrent linear branch that maintains a fixed-size state for streaming context.
Our key insight is that softmax attention captures fine-grained local interactions, while a recurrent state provides a controllable mechanism for long-range memory.
This design achieves linear-time scaling with constant memory while improving temporal consistency over the full-softmax model.
To prevent noisy intermediate states from corrupting long-range memory, we update the recurrent state only after the denoised pass. To avoid within-frame information asymmetry, all tokens share the same pre-update state rather than sequential updates.
To the best of our knowledge, this is the first work to convert a pretrained AR video diffusion model into a hybrid linear attention architecture, through an efficient two-stage training scheme adapted to the AR video setting.
With $75\%$ of layers replaced by hybrid linear attention, the model achieves up to $2.26\times$ wall-clock speedup and $54\%$ memory reduction, while maintaining comparable quality with improving temporal consistency.
\end{abstract}

\section{Introduction}
\label{sec:intro}

Autoregressive video diffusion has emerged as a powerful paradigm for streaming and interactive video synthesis, owing to its causal, chunk-wise formulation that naturally aligns with the temporal structure of video and enables streaming inference via key-value caching~\citep{chen2024diffusionforcing, huang2025selfforcing, zhu2026causalforcing, yin2025causvid, cui2025selfforcingpp, liu2026rollingforcing, liu2026diagonal, yang2025longlive, sand2025magi}.
However, all these systems employ softmax self-attention~\citep{vaswani2017attention} in Diffusion Transformers~\citep{peebles2023dit}, whose $O(N^{2})$ compute and $O(N)$ memory scaling in sequence length $N$ introduces a structural bottleneck that worsens with every generated block~\citep{xu2026kvcachesurvey, lv2026lightforcing}.
This cost is already severe at second-scale durations: the KV cache of a 5-second 480p video can exceed 34\,GB, surpassing the model parameters themselves~\citep{xi2026qvg}; and attention accounts for ${\sim}$75\% of total latency after only 14 chunks~\citep{lv2026lightforcing}, making it the dominant barrier to practical streaming deployment~\citep{yang2025longlive}.
\begin{figure}[t]
  \centering
  \includegraphics[width=\linewidth]{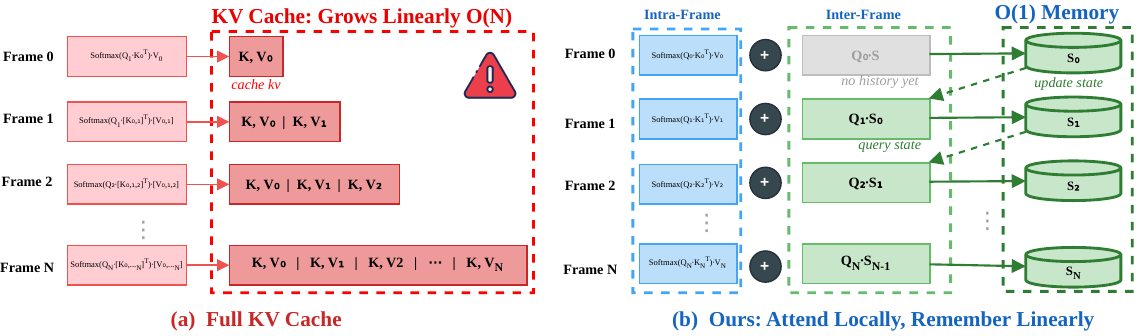}
  \caption{(a) AR video diffusion relies on softmax self-attention with a growing KV cache, incurring $O(N)$ memory, which limits scalability. (b) Replacing self-attention with our hybrid linear attention, the model maintains a fixed-size recurrent state for streaming context with $O(1)$ memory.}
  \vspace{-2em}
  \label{fig:intro-overview}
\end{figure}

Previous efforts have primarily explored sparse attention~\citep{lv2026lightforcing, guo2026dummyforcing, huang2025selfforcing, yang2025longlive, liu2026rollingforcing}, KV cache quantization and eviction~\citep{xi2026qvg, chen2026pafukv, yang2026tempcache, li2026packcache} to mitigate this cost. However, these methods have inherent limitations: sparse attention reduces per-step FLOPs but still relies on a linearly growing KV cache; quantization reduces per-entry precision but retains $O(N)$ storage; and eviction bounds memory usage at the cost of irreversible loss of previous context. Consequently, these methods do not simultaneously address the $O(N)$ memory scaling and streaming context management.

These limitations motivate rethinking the problem from a different modeling perspective. Recent work in language modeling reinterprets linear attention as a form of associative memory~\citep{schlag2021linear}, with modern variants introducing gating and error-correction mechanisms for more precise and stable updates~\citep{yang2024gla, yang2024deltanet, yang2025gdn}. 
Building on this perspective, recent approaches replace partial softmax attention layers with linear attention via distillation or retraining, improving long-context modeling and outperforming either paradigm alone~\citep{de2024griffin, chen2026halo}.

However, directly transferring these advances to AR video diffusion is non-trivial. While linear attention can serve as a recurrent memory for long-range context in LLMs, no prior work has explored how to convert a pretrained AR video diffusion model into such a hybrid architecture. Unlike the homogeneous causal attention used in LLMs, AR video diffusion exhibits an inherently heterogeneous attention structure: intra-frame bidirectional attention captures rich spatial correlations~\citep{fan2025rala}, while inter-frame causal attention relies on a growing KV cache to maintain temporal consistency~\citep{sand2025magi}.
This heterogeneity introduces two key challenges absent in the LLM setting: 
\textbf{(1)}~how to design a hybrid attention module that respects the distinct roles of intra- and inter-frame attention; \textbf{(2)}~how to maintain a stable recurrent state across frame-level diffusion denoising.

\begin{figure}[t]
  \centering
  \includegraphics[width=\linewidth]{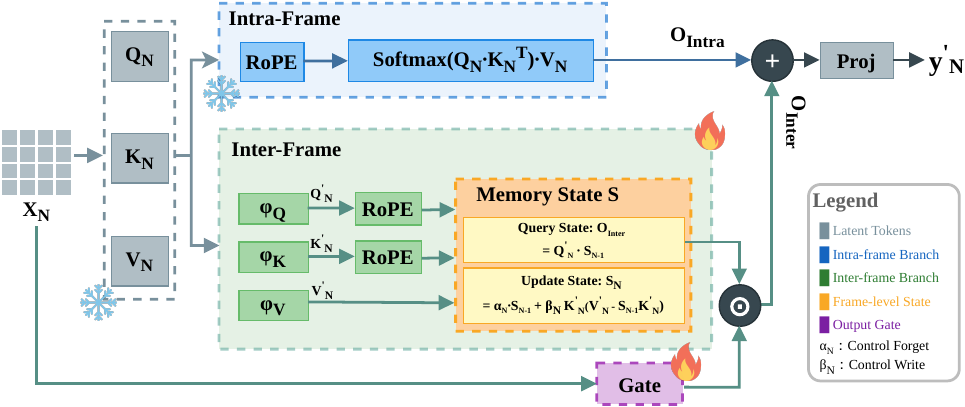}
  \caption{
ARL$^2$ attention module (normalization omitted for clarity). 
Given tokens $X_N$ of frame $N$, $Q_N$, $K_N$, $V_N$, which are routed to two branches. 
The intra-frame branch (top) applies bidirectional softmax attention over tokens within the current frame, producing $O_{\text{intra}}$. 
The inter-frame branch (bottom) applies recurrent linear attention, where a fixed-size state $S$ maintains and updates long-range memory across frames, and produces $O_{\text{inter}}$.
The state $S$ is updated once per frame. 
During training, the hybrid module is learned via a progressive two-stage distillation pipeline (See Sec.~\ref{sec:training-pipeline}).
}
  \vspace{-1.5em}
  \label{fig:method-overview}
\end{figure}

We address \raisebox{-1.1pt}{\ding[1.1]{182\relax}} challenge with a role-aware decomposition of attention into intra-frame and inter-frame branches. Inter-frame causal attention maintains temporal consistency over a growing context, while intra-frame bidirectional attention models fine-grained spatial structure (Fig.~\ref{fig:intro-overview}). Accordingly, we replace the KV cache with a fixed-size recurrent state via a Gated Delta Network~\citep{yang2025gdn} (\emph{remember linearly}), whose gated forgetting and delta update rules enable controllable long-range memory management at constant cost, while retaining intra-frame softmax attention to preserve spatial fidelity (\emph{attend locally}; Fig.~\ref{fig:method-overview}). Since all tokens within a frame are decoded jointly, retaining intra-frame softmax does not introduce additional latency.

The \raisebox{-1.1pt}{\ding[1.1]{183\relax}} challenge arises from the interaction between recurrent state and frame-level diffusion denoising. Sequential state updates would introduce intra-frame information asymmetry, and repeated exposure to noisy intermediates can corrupt the accumulated memory over long horizons. To address this, we introduce two mechanisms (Fig.~\ref{fig:state}): all tokens share a pre-update state to ensure consistent context, and the recurrent state is updated only after the final clean pass to prevent contamination. Ablations (Sec.~\ref{sec:experiments}) show that they are critical for maintaining video quality.

To realize this design in practice, we propose a progressive two-stage distillation pipeline to efficiently and stably convert a pretrained Causal Forcing model~\citep{zhu2026causalforcing} into the hybrid architecture.
Stage~1 performs per-layer distillation employing sensitivity-guided layer selection adapted to the AR video setting to minimize performance degradation, while Stage~2 conducts joint distillation to train the hybrid model end-to-end.
With 50\% of attention layers replaced by the hybrid linear attention, the hybrid model achieves up to $1.57\times$ wall-clock speedup and $35\%$ memory reduction, while maintaining comparable quality 
and improving temporal consistency with Temporal Flickering 
and Motion Smoothness 
, compared to the full-softmax teacher. At 75\% replacement, efficiency further improves to $2.26\times$ speedup and $54\%$ memory reduction.

\section{Related work}
\label{sec:related}

\paragraph{Autoregressive video diffusion.}
Representative open-source video Diffusion Transformers (DiTs), such as Wan~\citep{wan2025}, CogVideoX~\citep{yang2024cogvideox}, and Open-Sora~\citep{zheng2024opensora}, rely on spatiotemporal self-attention, leading to quadratic scaling in compute and memory with video length.
To enable streaming generation, a recent line of work reformulates video diffusion as frame-autoregressive (AR): Diffusion Forcing~\citep{chen2024diffusionforcing} assigns independent noise levels per token, bridging AR and full-sequence diffusion; Self-Forcing~\citep{huang2025selfforcing} closes the train-test exposure gap by conditioning on self-generated context during training, achieving real-time streaming; and Causal Forcing~\citep{zhu2026causalforcing} further improves quality through ODE distillation with an AR teacher. 
However, all these AR approaches are bottlenecked by linearly growing KV cache.
We propose a hybrid attention module for autoregressive video diffusion that replaces inter-frame softmax attention with a fixed-size recurrent state and retains intra-frame bidirectional softmax interactions, achieving $O(1)$ memory per-layer with respect to sequence length.

\paragraph{Linear attention for video.}
Existing efforts to incorporating linear attention into video generation can be broadly categorized as follows.
\textit{Full-architecture retraining.}
SANA-Video~\citep{chen2026sanavideo} adopts block-linear attention with constant-memory states, while SSM-based models explore alternative sequence modeling paradigms for long-horizon video diffusion and world models~\citep{oshima2024ssm, hong2025hth, po2025longctx}. Although effective, these methods require large-scale retraining. 
\textit{Supplementary branch.}
SALAD~\citep{fang2025salad} introduces a lightweight linear attention branch alongside sparse softmax attention, and VideoSSM~\citep{yu2025videossm} combines sliding-window softmax with an SSM-based global memory. In both cases, linear-complexity modules serve as auxiliary components, while the primary softmax path and its associated complexity remain.
\textit{Attention replacement in bidirectional DiTs.}
Attention Surgery~\citep{ghafoorian2025surgery} and SLA~\citep{zhang2026sla} replace selected attention blocks in pretrained video DiTs with kernel-based or sparse-linear approximations, achieving efficiency gains with minimal fine-tuning. However, they operate within the bidirectional denoising paradigm, where each step processes the full spatiotemporal sequence, so memory and computation still scale with video length and do not support streaming generation.
ReHyAt~\citep{ghafoorian2026rehyat} introduces a recurrent hybrid attention formulation for chunk-wise inference with constant memory, but is distilled from bidirectional diffusion models that assume full-sequence availability, leading to a mismatch for causal generation.
\textit{In contrast}, we convert a pretrained AR video diffusion model into a hybrid architecture with gated recurrent linear attention using a Casual Forcing teacher to preserve causal attention behavior. This is the first work to extend hybrid-attention distillation to the AR setting, enabling streaming video generation with constant memory without retraining from scratch.

\section{Method}
\label{sec:method}
We first review the attention mechanism in AR video diffusion and its memory bottleneck (Sec.~\ref{sec:preliminaries}). We then introduce ARL$^2$, a hybrid linear attention module that decomposes self-attention into intra-frame softmax and inter-frame recurrent branches (Sec.~\ref{sec:arl}). Next, we describe two state management strategies: frame-level access for consistent history observation and clean-pass-only updates to prevent contamination from noisy intermediates (Sec.~\ref{sec:state-management}). Finally, we present a lightweight two-stage distillation pipeline that converts a pretrained AR model into the hybrid architecture (Sec.~\ref{sec:training-pipeline}).

\subsection{Preliminaries}
\label{sec:preliminaries}
The autoregressive video diffusion model~\citep{chen2024diffusionforcing, huang2025selfforcing, zhu2026causalforcing} generates video chunk-by-chunk in an autoregressive manner, where each chunk contains one or more contiguous frames.
Let $x_N\!\in\!\mathbb{R}^{L\times d}$ denote the $L$ latent tokens of frame~$N$, where $d$ is the hidden dimension.
At each self-attention layer, the model computes queries, keys, and values $Q\!=\!x W_q$, $K\!=\!x W_k$, $V\!=\!x W_v$ for the current frame, and concatenates keys and values from previous frames into the KV cache.
The attention output over the full context is:
\begin{align}
\label{eq:softmax-attn}
y = \mathrm{softmax}\!\left(\frac{Q\, K^\top}{\sqrt{d_k}} + M \right) V,
\end{align}
where the causal block mask~$M$ permits bidirectional attention within the current frame and causal attention to previous frames.
As video length grows, the KV cache accumulates linearly: generating frame~$N$ requires storing $(N\!-\!1)\!\times\!L$ key-value pairs per layer, making memory scale as $\mathcal{O}(N)$.

\subsection{ARL$^2$: Attend Locally, Remember Linearly}
\label{sec:arl}
Our key insight stems from the heterogeneous attention structure of autoregressive video diffusion: intra-frame attention is bidirectional, modeling fine-grained spatial correlations, while inter-frame attention is causal and relies on a growing KV cache to maintain temporal consistency. Exploiting this asymmetry, we decompose the original self-attention into two branches (Fig.~\ref{fig:method-overview}).
One branch preserves the original intra-frame softmax attention, reusing the frozen pretrained projections:
\begin{align}
O_{\mathrm{intra}} &= \operatorname{Softmax}(\mathrm{RoPE}(Q), \mathrm{RoPE}(K), V),
\end{align}

The other branch replaces the inter-frame KV cache with a recurrent linear state $S\!\in\!\mathbb{R}^{H\times D\times D}$. Because the pretrained $Q,K,V$ are optimized for softmax attention, we introduce per-head grouped $1\!\times\!1$ feature maps $\phi_q,\phi_k,\phi_v$ to adapt the representations to the linear recurrence domain:
\begin{align}
\label{eq:feature-maps}
Q' &= \mathrm{L2Norm}(\mathrm{RoPE}(\phi_q(Q))), \quad
K' = \mathrm{L2Norm}(\mathrm{RoPE}(\phi_k(K))), \quad
V' = \phi_v(V),
\end{align}

We build our linear attention module on the Gated Delta Network~\citep{yang2025gdn}, whose state update rule is:
\begin{align}
\label{eq:gdn-update}
S_N = \alpha_N\, S_{N-1} + \beta_N\, {K'_N}^{\!\top}\!\left(V'_N - K'_N\, S_{N-1} \right),
\end{align}
where $\alpha_N$ (forget gate) and $\beta_N$ (learning rate) are produced by per-token projections of the hidden state. Eq.~\ref{eq:gdn-update} uses frame-level notation for brevity; in practice the update is applied sequentially over all $L$ tokens within the frame via an efficient chunkwise kernel~\citep{yang2024gla}, with per-token gates $\alpha^{(j)}, \beta^{(j)}$ for $j \in \{1,\dots,L\}$.
\begin{figure}[t]
\centering

\begin{subfigure}[t]{0.32\linewidth}
  \centering
  \includegraphics[width=\linewidth]{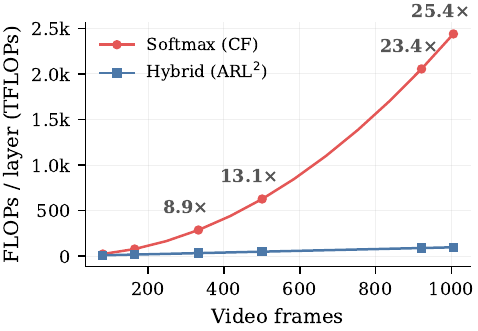}
  \caption{Theoretical per-layer FLOPs.}
  \label{fig:flops}
\end{subfigure}
\hfill
\begin{subfigure}[t]{0.32\linewidth}
  \centering
  \includegraphics[width=\linewidth]{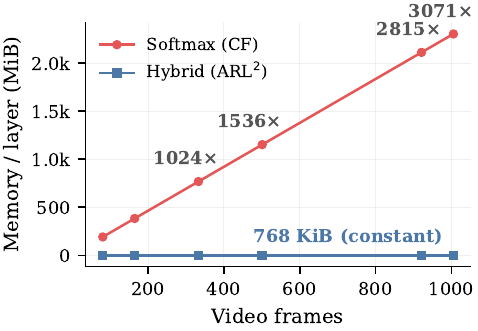}
  \caption{Per-layer memory scaling.}
  \label{fig:memory}
\end{subfigure}
\hfill
\begin{subfigure}[t]{0.32\linewidth}
  \centering
  \includegraphics[width=\linewidth]{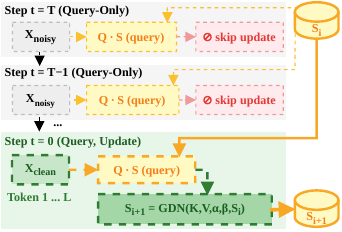}
  \caption{State management.}
  \label{fig:state}
\end{subfigure}

\caption{
(a) Softmax attention scales quadratically with video length, while the hybrid layer scales linearly.
(b) The KV cache grows linearly with video length, whereas ours remains constant.
(c) The recurrent state $S$ is shared across all tokens within a frame and updated only after the final clean pass.
}
\label{fig:flop&mem}
\vspace{-1.5em}
\end{figure}
When generating frame~$N$, the inter-frame output is obtained by querying the state conditioned on all previous frames:
\begin{align}
O_{\mathrm{inter}} = Q'_N\, S_{N-1},
\end{align}
where all tokens in the current frame query the same pre-update state $S_{N-1}$.
Then two branches are combined via a lightweight headwise gate:
\begin{align}
\label{eq:gate}
G = \sigma(x\, W_g), \qquad \\
y' = \left(O_{\mathrm{intra}} + G \odot O_{\mathrm{inter}} \right) W_o.
\end{align}
where $W_g\!\in\!\mathbb{R}^{d\times H}$ projects to one scalar per head, $\sigma$ denotes the sigmoid function, and $G\!\in\![0,1]^{L\times H}$ is broadcast over the head dimension $D$ before the element-wise product with $O_{\mathrm{inter}}\!\in\!\mathbb{R}^{L\times H\times D}$.

The fixed-size state $S\!\in\!\mathbb{R}^{H\times D\times D}$ reduces cumulative per-layer FLOPs from quadratic to linear scaling and keeps per-layer memory constant with respect to the number of generated frames (Fig.~\ref{fig:flops},~\ref{fig:memory}). In practice, replacing 75\% of layers yields up to $2.26\times$ wall-clock speedup and $54\%$ memory reduction (Sec.~\ref{sec:efficiency-scaling}).

\subsection{Memory state management for AR video diffusion}
\label{sec:state-management}

Recurrent linear attention in LLMs~\citep{yang2024gla, yang2024deltanet, yang2025gdn} updates the state at each token sequentially. Directly applying this to AR video diffusion introduces two issues: (1) tokens within a frame should share the same cross-frame context, but token-level updates cause asymmetry; (2) multi-step denoising writes noisy intermediates into the state, corrupting accumulated memory. We address both with two complementary mechanisms (Fig.~\ref{fig:state}).

\paragraph{Frame-level access.}
All $L$ tokens within a frame query the same pre-update state $S_{N-1}$, and the state is updated only after the entire frame has been processed:
\begin{align}
O_{\mathrm{inter},N}^{(i)} = Q'^{(i)}_N\, S_{N-1}, \quad \forall\, i \in \{1,\dots,L\}.
\end{align}

\paragraph{Clean-pass update.}
During the multi-step denoising loop, the model queries $S$ but does not update it. Once a clean frame is produced, its tokens update S via Eq.~\ref{eq:gdn-update}. This also reduces the number of state writes from one per denoising step to one per frame, improving inference efficiency. 


\subsection{Training pipeline: Two-stage distillation}
\label{sec:training-pipeline}
Rather than retraining the full model, we selectively replace a subset of attention layers and distill them in a progressive two-stage pipeline: Stage~1 independently distills each attention layer and select which layers to replace; Stage~2 jointly fine-tunes the hybrid model so that all replaced layers operate coherently. This yields strong performance with modest compute (Sec.~\ref{sec:experiments}), consistent with findings on progressive distillation for hybrid LLMs~\citep{chen2026halo, li2025klguided}.
\paragraph{Stage 1: per-layer distillation and selection.}
We distill all 30 attention layers independently. For each layer~$\ell$, the teacher attention and student hybrid module run in parallel on the teacher hidden state, and we minimize their output MSE:
\begin{align}
\label{eq:align-loss}
\mathcal{L}_{\mathrm{align}}^{(\ell)} =
\frac{1}{L \cdot d}\left\| y'_{\ell}(x_t, t, c) - y_{\ell}(x_t, t, c) \right\|_F^2,
\end{align}
where $y_\ell$ and $y'_\ell$ denote the teacher and student outputs, $x_t$ is the noisy latent at timestep~$t$, and $c$ is the text condition. The teacher output is propagated to subsequent layers, ensuring distillation under the intact teacher context. Only the state-related parameters are trained, including feature maps $\phi_q, \phi_k, \phi_v$ (Eq.~\ref{eq:feature-maps}), the headwise gate $W_g$ (Eq.~\ref{eq:gate}), and the state update projections for $\alpha_N, \beta_N$ (Eq.~\ref{eq:gdn-update}). Training uses text-conditioned inputs with randomly sampled noisy latents.
After per-layer distillation, we determine which layers to replace, in three steps: 
\begin{figure}[t]
\centering

\begin{subfigure}[t]{0.32\linewidth}
  \centering
  \includegraphics[width=\linewidth]{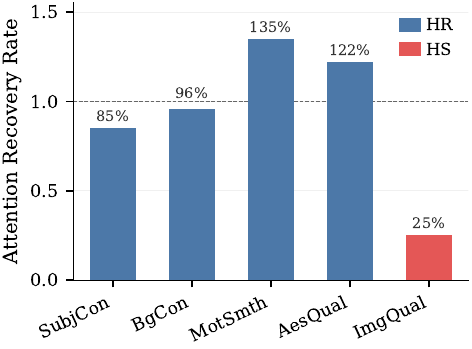}
  \caption{Per-dimension ARR analysis.}
  \label{fig:stage1-a}
\end{subfigure}
\hfill
\begin{subfigure}[t]{0.32\linewidth}
  \centering
  \includegraphics[width=\linewidth]{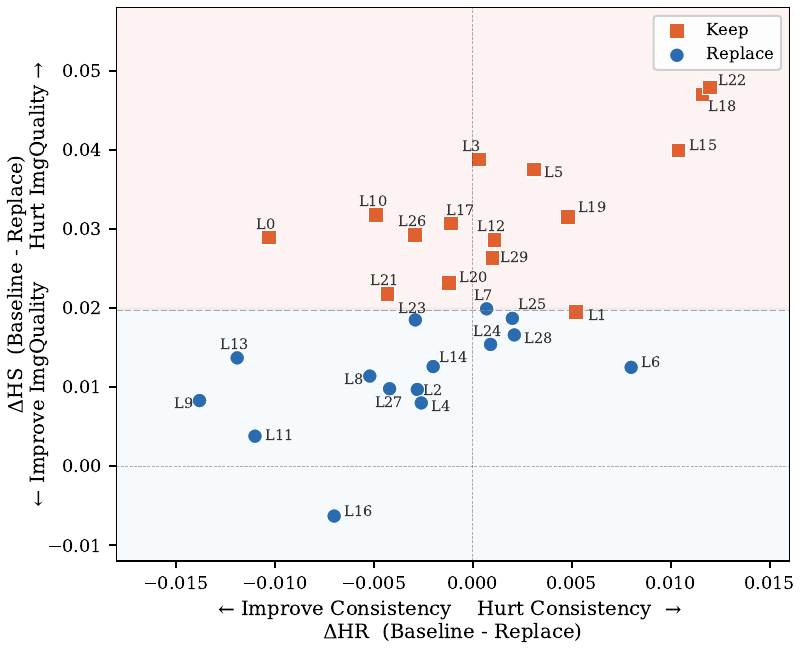}
  \caption{HS-protected layer selection.}
  \label{fig:stage1-b}
\end{subfigure}
\hfill
\begin{subfigure}[t]{0.32\linewidth}
  \centering
  \includegraphics[width=\linewidth]{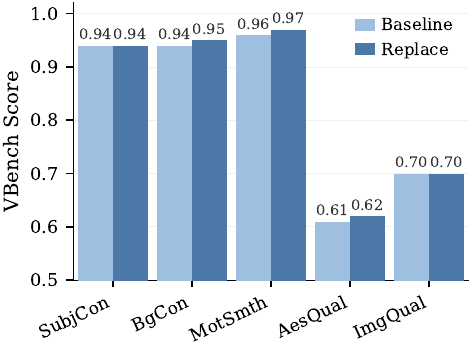}
  \caption{End-to-end validation.}
  \label{fig:stage1-c}
\end{subfigure}

\caption{
(a) Per-dimension ARR averaged across layers. Four Hybrid-Recoverable (HR) dimensions achieve $\ge$85\% recovery, while ImgQual is Hybrid-Sensitive (HS) at $\sim$25\%, motivating HS-aware layer selection to minimize quality degradation.
(b) Each point corresponds to one of the 30 layers, plotted by $\Delta\mathrm{HR}$ (x-axis) and $\Delta\mathrm{HS}$ (y-axis). Here the 50\% lowest-$p_\ell$ layers (blue) are selected for replacement.
(c) End-to-end validation after Stage~2: the hybrid model matches or exceeds the baseline on all HR dimensions and recovers ImgQual to within $\sim$1\%, indicating that HS-aware selection preserves sensitive dimensions while allowing HR gaps to be closed via joint distillation.
}

\label{fig:stage1}
\vspace{-1.5em}
\end{figure}

\textbf{(i) Layer analysis.}
For each layer we evaluate three conditions on our stress-test prompts and measure VBench scores across 5 quality dimensions (indexed by $i$):
(1)~\textbf{Baseline}$_i$: original softmax with full KV cache;
(2)~\textbf{Replace}$_{\ell,i}$: layer~$\ell$ alone replaced with its Stage-1 hybrid module;
(3)~\textbf{Skip}$_{\ell,i}$: attention output of layer~$\ell$ zeroed via a modulated skip connection.
From these VBench scores we compute the Attention Recovery Rate (ARR) per layer~$\ell$ and quality dimension~$i$:
\begin{align}
\mathrm{ARR}_{\ell,i} = \frac{\text{Replace}_{\ell,i} - \text{Skip}_{\ell,i}}{\text{Baseline}_{i} - \text{Skip}_{\ell,i}},
\label{eq:arr}
\end{align}
where $\mathrm{ARR}{=}1$ denotes full recovery of the softmax attention's contribution to dimension~$i$. 
Aggregating ARR across layers (Fig.~\ref{fig:stage1-a} splits the five dimensions into two groups. \textbf{Hybrid-Recoverable (HR)}: SubjCon, BgCon, MotSmth, AesQual, where hybrid module achieves $\ge$85\% recovery. \textbf{Hybrid-Sensitive (HS)}: ImgQual, where hybrid module recovers only $\sim$25\% of the attention contribution. 

\textbf{(ii) Layer selection.}
Motivated by the hypothesis that HS dimensions are intrinsically sensitive to attention replacement, we design layer selection to explicitly safeguard them. Concretely, each layer is scored by its impact on HS, with a small HR penalty acting as a tiebreaker:
\begin{align}
p_\ell \;=\; {\Delta\mathrm{HS}_\ell}
\;+\; \beta \cdot \max\!\bigl(\Delta\mathrm{HR}_\ell,\,0\bigr),
\label{eq:protection-score}
\end{align}
where $\Delta\mathrm{HS}_\ell$ and $\Delta\mathrm{HR}_\ell$ denote the mean degradation across HS and HR dimensions, respectively. Under a 50\% layer replacement setting, we replace the 15 layers with the smallest $p_\ell$ (Fig.~\ref{fig:stage1-b}). 
Sweeping $\beta\!\in\!\{0.1,0.5,1.0\}$ yields the identical replacement set, so $\beta$ functions as a boundary-layer tiebreaker rather than a sensitive hyperparameter. 

\textbf{(iii) End-to-end validation.}
After Stage~2 joint distillation, the resulting hybrid model matches or exceeds the softmax baseline on all four HR dimensions and recovers ImgQual to within $\sim$1\% on the same stress-test prompts (Fig.~\ref{fig:stage1-c}). These results indicate that the HS-protected layer selection effectively prevents severe degradation in hybrid-sensitive layers, while allowing residual errors in hybrid-recoverable layers to be compensated during joint training. 

\paragraph{Stage 2: joint distillation.}
The student is initialized from the teacher checkpoint, the Stage~1 GDN weights are loaded into the selected layers, and the resulting hybrid model is jointly distilled by matching the teacher's predicted velocity field. Let $v_\theta(x_t, t, c)$ denote the flow velocity predicted by the diffusion model given noisy latent $x_t$, timestep $t$, and text condition $c$. The joint loss is:
\begin{align}
\mathcal{L}_{\mathrm{joint}} =
\left\| v_{\theta}^{\mathrm{student}}(x_t, t, c) -
v_{\theta}^{\mathrm{teacher}}(x_t, t, c) \right\|_2^2 .
\end{align}
In this stage, the teacher is frozen, while the student trains the hybrid linear attention module parameters, LoRA adapters on the Q, K, V path, and the FFN of each selected block.

\begin{table*}[t]
  \vspace{-8pt}
  \centering
  \footnotesize
  \setlength{\tabcolsep}{3pt}
  \renewcommand{\arraystretch}{1.25}
  \caption{ARL$^2$ achieves comparable overall performance to representative open-source video models, while improving temporal flickering and background consistency, with $\mathcal{O}(1)$ memory complexity. ARL$^2$ (50\%/75\%) denotes replacing 50\%/75\% of attention layers with hybrid linear attention.}
  \label{tab:vbench-baselines}
  \resizebox{\textwidth}{!}{%
  \begin{tabular}{l c *{6}{c} c c}
    \toprule
    \raisebox{0.8ex}{\textbf{Model}}
      & \raisebox{0.8ex}{\textbf{\#Params}}
      & \shortstack{Subject\\Consistency}
      & \shortstack{Background\\Consistency}
      & \shortstack{Temporal\\Flickering}
      & \shortstack{Motion\\Smoothness}
      & \shortstack{Aesthetic\\Quality}
      & \shortstack{Imaging\\Quality}
      & \shortstack{\textbf{Quality}\\\textbf{Avg.}$\uparrow$}
      & \shortstack{\textbf{Semantic}\\\textbf{Avg.}$\uparrow$} \\
    \midrule
    \rowcolor{paradigmgray}
    \multicolumn{10}{c}{\textbf{Bidirectional Diffusion}} \\
    \quad LTX-Video~\citep{hacohen2024ltx}  & 1.9B & 87.89 & 92.00 & 98.20 & 97.91 & 48.08 & 58.62 & 80.45 & 51.32 \\
    \quad Wan2.1-T2V~\citep{wan2025} & 1.3B & 96.18 & 98.33 & 98.33 & 96.60 & 65.35 & 68.64 & 87.24 & 63.01 \\
    \rowcolor{paradigmgray}
    \multicolumn{10}{c}{\textbf{Linear Diffusion}} \\
    \quad SANA-Video~\citep{chen2026sanavideo}       & 2B & 97.09 & 96.95 & 89.22 & 95.10 & 70.48 & 67.55 & 86.07 & \textbf{80.76} \\
    \rowcolor{paradigmgray}
    \multicolumn{10}{c}{\textbf{Autoregressive Diffusion}} \\
    \quad SkyReels-V2~\citep{chen2025skyreels}$^{\dagger}$ & 1.3B & 97.78 & 97.68 & 97.97 & 98.50 & 59.66 & 62.29 & 85.65 & 48.02 \\
    \quad MAGI-1~\citep{sand2025magi}       & 4.5B & 95.71 & 97.14 & 97.69 & 98.67 & 62.44 & 64.84 & 86.08 & 72.02 \\
    \rowcolor{paradigmgray}
    \multicolumn{10}{c}{\textbf{Distilled Autoregressive}} \\
    \quad Self Forcing~\citep{huang2025selfforcing}    & 1.3B & 95.62          & 96.40          & 96.82          & \textbf{95.59} & 66.75          & 69.96          & 86.86          & 70.71          \\
    \quad Causal Forcing~\citep{zhu2026causalforcing}  & 1.3B & \textbf{96.19} & \textbf{96.44} & 96.27          & 94.24          & \textbf{68.80} & \textbf{70.62} & 87.09          & 70.97          \\
    \rowcolor{paradigmgray}
    \multicolumn{10}{c}{\shortstack[l]{\textbf{Distilled Linear Autoregressive}}} \\
    \rowcolor{oursgreen}
    \quad ARL$^2$ (50\%)  & 1.3B & 95.63          & 96.41          & \textbf{97.26} & \textbf{95.59} & 68.29          & 69.87          & \textbf{87.17} & 70.13          \\
    \rowcolor{oursgreen}
    \quad ARL$^2$ (75\%)  & 1.3B & 94.85          & 95.94          & 96.90          & 94.24          & 68.44          & 69.87          & 86.71          & 70.24          \\
    \bottomrule
  \end{tabular}%
  }
  \vspace{-12pt}
\end{table*}

\section{Experiments}
\label{sec:experiments}

\subsection{Setup}
\label{sec:exp-setup}

\paragraph{Training.} We adopt Causal Forcing~\citep{zhu2026causalforcing} as base model. In Stage~1, all 30 layers are independently distilled for 6K steps using prompts from VidProM~\citep{wang2024vidprom}; 
then identifies 15 layers (50\%) or 23 layers (75\%) for replacement. Stage~2 jointly distills the hybrid student against the frozen Causal Forcing teacher for 12K steps on 6K teacher-synthesized videos.  The entire pipeline trains fewer than 2\% of backbone parameters, consumes ${\sim}$156 H100 GPU-hours (${\sim}$50 for Stage~1, ${\sim}$106 for Stage~2). 

\paragraph{Evaluation.}
We use VBench~\citep{huang2024vbench} as the primary benchmark. All models are evaluated at $832\times480$ with 81 frames ($\dagger$: native resolution $544\times960$, 97 frames). Layer sensitivity analysis uses a separate set of stress-test prompts designed to probe temporal consistency and visual fidelity under challenging conditions (details in App.~\ref{app:evaluation}). For efficiency, we evaluate end-to-end generation time, throughput (FPS), and peak GPU memory on a subset of prompts, measured on a single
\begin{wraptable}{r}{0.6\textwidth}
  \vspace{-10pt}
  \centering
  \scriptsize
  \setlength{\tabcolsep}{3pt}
  \renewcommand{\arraystretch}{1.3}
  \setlength{\intextsep}{4pt}
  \setlength{\columnsep}{6pt}
  \caption{ARL$^2$ achieves consistent efficiency gains with up to $2.26\times$ speedup and significantly reduced memory, while maintaining competitive or improved video quality.}
  \begin{tabular}{@{}l @{\hspace{6pt}} l l l l l l l l l@{}}
    \toprule
    & & \multicolumn{4}{c}{\emph{Efficiency}} & & \multicolumn{3}{c}{\emph{Quality}$\uparrow$} \\
    \cmidrule(lr){3-6} \cmidrule(lr){8-10}
    \shortstack[l]{Video\\Frames} & \raisebox{0.8ex}{Model} & \shortstack[l]{Gen\\(s)$\downarrow$} & \shortstack[l]{FPS\\$\uparrow$} & \shortstack[l]{Mem\\(GB)$\downarrow$} & \shortstack[l]{Speed-\\up$\uparrow$} & & \shortstack[l]{Qual.\\Avg} & \shortstack[l]{Sem.\\Avg} & \shortstack[l]{~\\Total} \\
    \midrule
    \multirow{4}{*}{\makebox[2.2em][l]{81}}
      & Self Forcing     & 8.70  & 9.31 & 23.30 & 1.00$\times$ & & 86.79 & 69.15 & 83.26 \\
      & Causal Forcing   & 8.68  & 9.33 & 23.30 & 1.00$\times$ & & 86.69 & 69.11 & 83.17 \\
      & \gc ARL$^2$ (50\%)   & \gc 8.70  & \gc 9.31 & \gc 20.53 & \gc 1.00$\times$ & \gc & \gc \textbf{87.81} & \gc \textbf{71.41} & \gc \textbf{84.53} \\
      & \gc ARL$^2$ (75\%)   & \gc \textbf{8.67}  & \gc \textbf{9.34} & \gc \textbf{19.10} & \gc 1.00$\times$ & \gc & \gc 86.00 & \gc 65.86 & \gc 81.97 \\
    \midrule
    \multirow{4}{*}{\makebox[2.2em][l]{165}}
      & Self Forcing     & 22.27 & 7.41 & 29.13 & 1.00$\times$ & & 86.01 & 66.25 & 82.05 \\
      & Causal Forcing   & 22.28 & 7.41 & 29.13 & 1.00$\times$ & & 85.48 & 70.55 & 82.49 \\
      & \gc ARL$^2$ (50\%)   & \gc 19.90 & \gc 8.29 & \gc 23.54 & \gc 1.12$\times$ & \gc & \gc \textbf{86.32} & \gc 68.65 & \gc \textbf{82.79} \\
      & \gc ARL$^2$ (75\%)   & \gc \textbf{18.63} & \gc \textbf{8.86} & \gc \textbf{20.60} & \gc \textbf{1.20}$\times$ & \gc & \gc 84.36 & \gc \textbf{71.75} & \gc 81.84 \\
    \midrule
    \multirow{4}{*}{\makebox[2.2em][l]{501}}
      & Self Forcing     & 124.24 & 4.03 & 54.08 & 1.00$\times$ & & 85.14 & 53.94 & 78.90 \\
      & Causal Forcing   & 124.24 & 4.03 & 54.08 & 1.00$\times$ & & 83.13 & \textbf{58.96} & 78.30 \\
      & \gc ARL$^2$ (50\%)   & \gc 88.52 & \gc 5.66 & \gc 37.26 & \gc 1.40$\times$ & \gc & \gc \textbf{86.04} & \gc 58.85 & \gc \textbf{80.60} \\
      & \gc ARL$^2$ (75\%)   & \gc \textbf{71.54} & \gc \textbf{7.00} & \gc \textbf{28.30} & \gc \textbf{1.74}$\times$ & \gc & \gc 77.00 & \gc 55.35 & \gc 72.67 \\
    \midrule
    \multirow{4}{*}{\makebox[2.2em][l]{921}}
      & Self Forcing     & 357.74 & 2.57 & 87.88 & 1.00$\times$ & & 83.37 & \textbf{67.83} & \textbf{80.26} \\
      & Causal Forcing   & 357.52 & 2.58 & 87.88 & 1.00$\times$ & & 79.96 & 58.43 & 75.66 \\
      & \gc ARL$^2$ (50\%)   & \gc 227.48 & \gc 4.05 & \gc 57.00 & \gc 1.57$\times$ & \gc & \gc \textbf{84.96} & \gc 51.62 & \gc 78.29 \\
      & \gc ARL$^2$ (75\%)   & \gc \textbf{158.16} & \gc \textbf{5.82} & \gc \textbf{40.50} & \gc \textbf{2.26}$\times$ & \gc & \gc 77.34 & \gc 48.40 & \gc 71.55 \\
    \bottomrule
  \end{tabular}
  \label{tab:efficiency-quality}
  \vspace{-40pt}
\end{wraptable}
NVIDIA RTX PRO 6000 Blackwell under cold-start conditions, across video lengths from 81 to 921 frames. 

\subsection{Results}
\label{sec:baselines}
\paragraph{Quality.}
As shown in Tab.~\ref{tab:vbench-baselines}, we compare ARL$^2$ with representative open-source video generation models, including LTX-Video~\citep{hacohen2024ltx}, Wan2.1~\citep{wan2025}, SANA-Video~\citep{chen2026sanavideo}, SkyReels-V2~\citep{chen2025skyreels}, MAGI-1~\citep{sand2025magi}), and Self Forcing~\citep{huang2025selfforcing}, Causal Forcing~\citep{zhu2026causalforcing}).

ARL$^2$ (50\%) achieves the higher Quality score among all distilled models ($87.17$), with improvements in temporal flickering ($+0.99$) and motion smoothness ($+1.35$), suggesting improved consistency.
ARL$^2$ achieves comparable quality to the retrained SANA-Video (linear model), while matching the much larger MAGI-1 ($4.5$B, $86.08$) with $3.5\times$ fewer parameters.
Details (App.~\ref{app:full-vbench}) show gains in object recognition ($+1.10$) and color fidelity ($+0.76$), but weaker performance on spatial relationships and human actions, suggesting limits of a fixed-size state for precise spatial modeling.

\paragraph{Efficiency.}
\label{sec:efficiency-scaling}

Tab.~\ref{tab:efficiency-quality} highlights our computational and memory efficiency through wall-clock time and peak memory across increasing video lengths. Speedup increases with video length, consistent with the linear–quadratic scaling gap in Fig.~\ref{fig:flops}. ARL$^2$ (75\%) reaches $1.20\times$, $1.74\times$, and $2.26\times$ at 165, 501, and 921 frames, respectively. Even ARL$^2$ (50\%) achieves $1.57\times$ at 921 frames, confirming that gains grow as KV-cache overhead dominates. 

For memory, hybrid layers replace the $O(N)$ KV-cache with a fixed-size state, reducing per-frame memory from $293$,MB to $175$,MB ($40\%$). At 921 frames, ARL$^2$ (75\%) uses $40.5$,GB vs.\ $87.9$,GB for Casual Forcing ($54\%$). At 1005 frames, Casual Forcing run out of memory ($>91$,GB), while ARL$^2$ completes in 179,s using 43,GB, demonstrating scalability. All results use batch size 1; the memory savings enable higher parallelism and throughput. 
Beyond efficiency, ARL$^2$ (50\%) maintains strong quality as video length increases. This suggests the recurrent state provides more stable long-range context than the KV-cache, consistent with improved inter-frame consistency in Tab.~\ref{tab:vbench-baselines}.

\begin{wrapfigure}{r}{0.6\linewidth}
  \vspace{-12pt}
  \centering
  \begin{subfigure}[t]{0.48\linewidth}
    \centering
    \includegraphics[width=\linewidth]{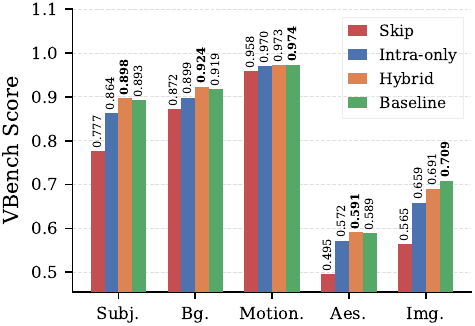}
    \caption{Attention decomposition.}
    \label{fig:attn-decomp}
  \end{subfigure}\hfill
  \begin{subfigure}[t]{0.48\linewidth}
    \centering
    \includegraphics[width=\linewidth]{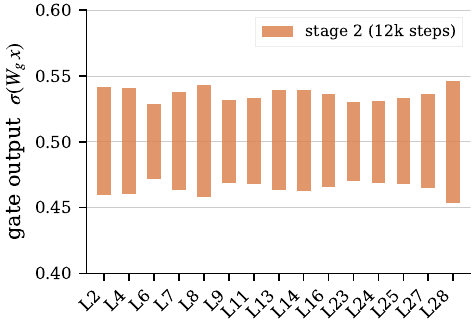}
    \caption{Gate activation.}
    \label{fig:gate-active}
  \end{subfigure}
  \vspace{-4pt}
  \caption{(a) Attention decomposition on a representative layer (23), highlighting complementary intra- and inter-frame branch modeling. (b) Gate activation across hybrid layers after Stage~2 training, indicating that both branches remain active and are combined via a content-dependent mixture.}
  \label{fig:ablation-vis}
  \vspace{-12pt}
\end{wrapfigure}

\subsection{Ablation study}
\label{sec:ablation}

We ablate two core design choices using 6 hybrid layers $[2,4,7,8,23,24]$ trained with Stage~1 and Stage~2 ($3$K steps each), evaluated on stress-test prompts at 21 frames.

\textbf{Branch contribution.}
To isolate intra- and inter-frame roles, we evaluate four settings on a single hybrid layer: \emph{Skip}, \emph{Intra-only}, \emph{Hybrid}, and \emph{Baseline}. As shown in Fig.~\ref{fig:attn-decomp} (layer 23 as an example), removing the layer (\emph{Skip}) degrades all metrics, indicating replaced layers remain essential. The intra-frame branch alone ((\emph{Intra-only})) largely recovers subject and background consistency, highlighting its role in local spatial reasoning. Adding the recurrent branch (\emph{Hybrid}) further improves background consistency and aesthetics, and slightly surpasses the teacher (\emph{Baseline}) on motion smoothness, suggesting complementary temporal modeling.
Fig.~\ref{fig:gate-active} shows the learned gate $\sigma(W_g x)$ stays in $[0.45, 0.55]$ with per-head variation, indicating both branches remain active and form a content-dependent mixture.

\textbf{State access granularity.}
We compare frame-level access (all tokens query the same $S_{N-1}$)
\begin{wraptable}{r}{0.55\linewidth}
  \vspace{-8pt}
  \centering
  \scriptsize
  \setlength{\tabcolsep}{2.5pt}
  \caption{Frame-level is superior to token-level access.}
  \begin{tabular}{lcccccc}
    \toprule
    Setting & SubjCon & BgCon & MotSmth & AesQual & ImgQual & Avg \\
    \midrule
    Causal Forcing & \textbf{93.7} & 93.7 & 96.5 & \textbf{61.3} & \textbf{70.9} & 83.2 \\
    Token-Level & 93.2 & 94.2 & 96.7 & 60.6 & 69.0 & 82.7 \\
    \rowcolor{vbenchblue}
    \textbf{Frame-Level} & 93.6 & \textbf{94.3} & \textbf{97.4} & 61.1 & 70.3 & \textbf{83.3} \\
    \bottomrule
  \end{tabular}
  \label{tab:access-ablation}
  \vspace{-12pt}
\end{wraptable}
against token-level access (each token sees its own sequentially accumulated state). Frame-level reaches Avg $83.3$, matching the full-softmax teacher ($83.2$), while token-level drops to $82.7$ (Tab.~\ref{tab:access-ablation}). This confirms that frame-level access is necessary to avoid intra-frame context asymmetry introduced by token-level updates.

\textbf{Gate granularity.}
The branch-fusion gate $\sigma(W_g x)$ can be parameterized as scalar ($1.5$K params per layer), headwise ($18$K), or elementwise ($2.36$M). Elementwise gating is  
\begin{wraptable}{r}{0.55\linewidth}
  \vspace{-8pt}
  \centering
  \scriptsize
  \setlength{\tabcolsep}{2.5pt}
  \caption{Headwise gating achieves the best trade-off between performance and parameter efficiency.}
  \begin{tabular}{lrcccccc}
    \toprule
    Setting & Params & SubjCon & BgCon & MotSmth & AesQual & ImgQual & Avg \\
    \midrule
    Elementwise & 2.36M & 92.6 & 93.8 & 96.1 & 60.0 & 67.8 & 82.1 \\
    \rowcolor{vbenchblue}
    Headwise & 18K & \textbf{93.8} & \textbf{94.5} & \textbf{97.6} & 61.3 & \textbf{70.3} & \textbf{83.5} \\
    Scalar & 1.5K & \textbf{93.8} & 94.4 & 97.5 & \textbf{61.4} & 69.6 & 83.3 \\
    \bottomrule
  \end{tabular}
  \label{tab:gate-ablation}
  \vspace{-18pt}
\end{wraptable}
strictly worse ($-1.4$), due to poor convergence under its $130\times$ larger parameterization with limited training steps. Scalar and headwise both match or exceed the casual forcing teacher, with headwise achieving the highest Avg ($83.5$) while providing per-head adaptability at modest cost.
\begin{figure}[t]
  \centering
  \includegraphics[width=\linewidth]{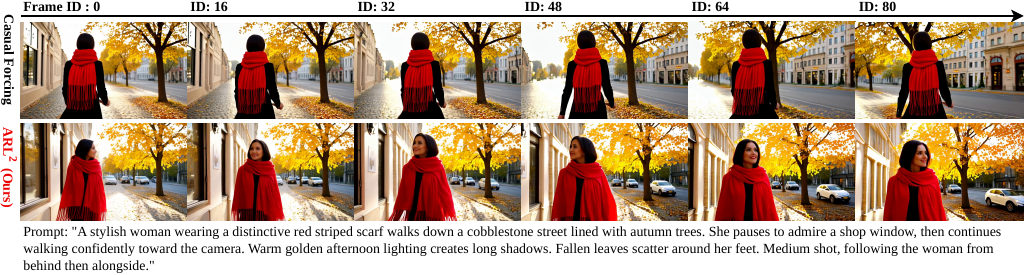}%
  \vspace{0.2em}
  \includegraphics[width=\linewidth]{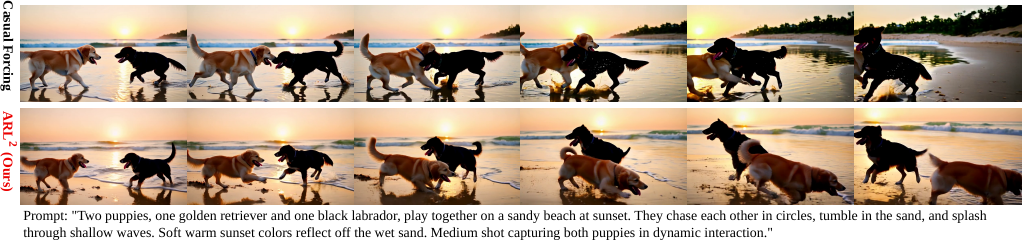}%
  \vspace{0.2em}
  \includegraphics[width=\linewidth]{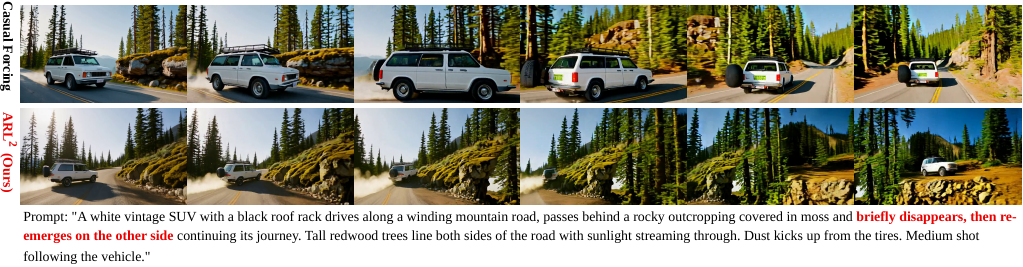}%
  \vspace{0.2em}
  \includegraphics[width=\linewidth]{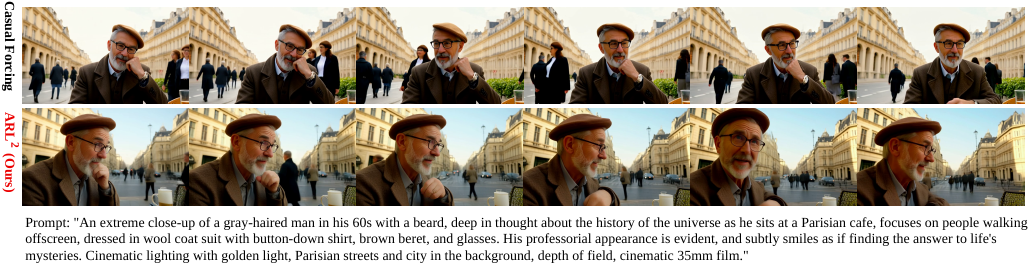}%
  \caption{Qualitative comparison. Highlighting strong visual quality and temporal consistency under significant computational and memory efficiency. ARL$^2$ achieves up to $2.26\times$ speedup and $54\%$ memory reduction with comparable or better quality. In the “white vintage SUV” example, ARL$^2$ preserves appearance consistency when the car disappears and reappears, while Causal Forcing fails to follow the prompt and produces distortions.}
  \label{fig:vis}
\end{figure}

\section{Conclusion}
\label{sec:conclusion}

We presented ARL$^2$, a hybrid attention architecture for autoregressive video diffusion that preserves bidirectional intra-frame softmax attention for fine-grained intra-frame spatial modeling while replacing the growing cross-frame KV cache with a fixed-size gated recurrent state. Two complementary mechanisms, frame-level state access and clean-pass updates, ensure stable streaming memory under iterative diffusion denoising. A lightweight two-stage distillation pipeline, guided by sensitivity-aware layer selection, converts a pretrained softmax model into the hybrid architecture with fewer than 2\% trainable parameters and ${\sim}$156 H100 GPU-hours.
At 50\% layer replacement, ARL$^2$ matches or exceeds the softmax teacher in overall quality while improving temporal consistency, with $1.57\times$ speedup and $35\%$ memory reduction. At 75\% replacement, efficiency scales to $2.26\times$ speedup and $54\%$ memory savings while maintaining competitive generation quality, enabling long-form video synthesis beyond the memory limits of full-attention baselines.



\bibliographystyle{unsrtnat}
\bibliography{ref}

@article{peebles2023dit,
  title={Scalable Diffusion Models with Transformers},
  author={William S. Peebles and Saining Xie},
  journal={2023 IEEE/CVF International Conference on Computer Vision (ICCV)},
  year={2022},
  pages={4172-4182},
  url={https://api.semanticscholar.org/CorpusID:254854389}
}

@article{wan2025,
  title={Wan: Open and Advanced Large-Scale Video Generative Models},
  author={Ang Wang and Baole Ai and Bin Wen and Chaojie Mao and Chen-Wei Xie and Di Chen and Feiwu Yu and Haiming Zhao and Jianxiao Yang and Jianyuan Zeng and Jiayu Wang and Jingfeng Zhang and Jingren Zhou and Jinkai Wang and Jixuan Chen and Kai Zhu and Kang Zhao and Keyu Yan and Lianghua Huang and Xiaofeng Meng and Ningying Zhang and Pandeng Li and Ping Wu and Ruihang Chu and Rui Feng and Shiwei Zhang and Siyang Sun and Tao Fang and Tianxing Wang and Tianyi Gui and Tingyu Weng and Tong Shen and Wei Lin and Wei Wang and Wei Wang and Wen-Chao Zhou and Wente Wang and Wen Shen and Wenyuan Yu and Xianzhong Shi and Xiaomin Huang and Xin Xu and Yan Kou and Yan-Mei Lv and Yifei Li and Yijing Liu and Yiming Wang and Yingya Zhang and Yitong Huang and Yong Li and You Wu and Yu Liu and Yulin Pan and Yun Zheng and Yuntao Hong and Yupeng Shi and Yutong Feng and Zeyinzi Jiang and Zhen Han and Zhigang Wu and Ziyu Liu},
  journal={ArXiv},
  year={2025},
  volume={abs/2503.20314},
  url={https://api.semanticscholar.org/CorpusID:277321639}
}

@article{yang2024cogvideox,
  title={CogVideoX: Text-to-Video Diffusion Models with An Expert Transformer},
  author={Zhuoyi Yang and Jiayan Teng and Wendi Zheng and Ming Ding and Shiyu Huang and Jiazheng Xu and Yuanming Yang and Wenyi Hong and Xiaohan Zhang and Guanyu Feng and Da Yin and Xiaotao Gu and Yuxuan Zhang and Weihan Wang and Yean Cheng and Ting Liu and Bin Xu and Yuxiao Dong and Jie Tang},
  journal={ArXiv},
  year={2024},
  volume={abs/2408.06072},
  url={https://api.semanticscholar.org/CorpusID:271855655}
}

@article{zheng2024opensora,
  title={Open-Sora: Democratizing Efficient Video Production for All},
  author={Zangwei Zheng and Xiangyu Peng and Tianji Yang and Chenhui Shen and Shenggui Li and Hongxin Liu and Yukun Zhou and Tianyi Li and Yang You},
  journal={ArXiv},
  year={2024},
  volume={abs/2412.20404},
  url={https://api.semanticscholar.org/CorpusID:275133398}
}

@article{chen2024diffusionforcing,
  title={Diffusion Forcing: Next-token Prediction Meets Full-Sequence Diffusion},
  author={Boyuan Chen and Diego Marti Monso and Yilun Du and Max Simchowitz and Russ Tedrake and Vincent Sitzmann},
  journal={ArXiv},
  year={2024},
  volume={abs/2407.01392},
  url={https://api.semanticscholar.org/CorpusID:270869622}
}

@article{huang2025selfforcing,
  title={Self Forcing: Bridging the Train-Test Gap in Autoregressive Video Diffusion},
  author={Xun Huang and Zhengqi Li and Guande He and Mingyuan Zhou and Eli Shechtman},
  journal={ArXiv},
  year={2025},
  volume={abs/2506.08009},
  url={https://api.semanticscholar.org/CorpusID:279251392}
}

@article{zhu2026causalforcing,
  title={Causal Forcing: Autoregressive Diffusion Distillation Done Right for High-Quality Real-Time Interactive Video Generation},
  author={Hongzhou Zhu and Min Zhao and Guande He and Hang Su and Chongxuan Li and Jun Zhu},
  journal={ArXiv},
  year={2026},
  volume={abs/2602.02214},
  url={https://api.semanticscholar.org/CorpusID:285269490}
}

@article{liu2026rollingforcing,
  title={Rolling Forcing: Autoregressive Long Video Diffusion in Real Time},
  author={Kunhao Liu and Wenbo Hu and Jiale Xu and Ying Shan and Shijian Lu},
  journal={ArXiv},
  year={2025},
  volume={abs/2509.25161},
  url={https://api.semanticscholar.org/CorpusID:281676207}
}

@article{cui2025selfforcingpp,
  title={Self-Forcing++: Towards Minute-Scale High-Quality Video Generation},
  author={Justin Cui and Jie Wu and Ming Li and Tao Yang and Xiaojie Li and Rui Wang and Andrew Bai and Yuanhao Ban and Cho-jui Hsieh},
  journal={ArXiv},
  year={2025},
  volume={abs/2510.02283},
  url={https://api.semanticscholar.org/CorpusID:281724666}
}

@inproceedings{liu2026diagonal,
  title={Streaming Autoregressive Video Generation via Diagonal Distillation},
  author={Jinxiu Liu and Xuan Liu and Kangfu Mei and Yandong Wen and Ming-Hsuan Yang and Weiyang Liu},
  year={2026},
  url={https://api.semanticscholar.org/CorpusID:286428940}
}

@inproceedings{schlag2021linear,
  title={Linear Transformers Are Secretly Fast Weight Programmers},
  author={Imanol Schlag and Kazuki Irie and J{\"u}rgen Schmidhuber},
  booktitle={International Conference on Machine Learning},
  year={2021},
  url={https://api.semanticscholar.org/CorpusID:235377069}
}

@article{de2024griffin,
  title={Griffin: Mixing Gated Linear Recurrences with Local Attention for Efficient Language Models},
  author={Soham De and Samuel L. Smith and Anushan Fernando and Aleksandar Botev and George Cristian-Muraru and Albert Gu and Ruba Haroun and Leonard Berrada and Yutian Chen and Srivatsan Srinivasan and Guillaume Desjardins and Arnaud Doucet and David Budden and Yee Whye Teh and Razvan Pascanu and Nando de Freitas and Caglar Gulcehre},
  journal={ArXiv},
  year={2024},
  volume={abs/2402.19427},
  url={https://api.semanticscholar.org/CorpusID:268091246}
}

@article{yang2024gla,
  title={Gated Linear Attention Transformers with Hardware-Efficient Training},
  author={Songlin Yang and Bailin Wang and Yikang Shen and Rameswar Panda and Yoon Kim},
  journal={ArXiv},
  year={2023},
  volume={abs/2312.06635},
  url={https://api.semanticscholar.org/CorpusID:266162792}
}

@article{yang2024deltanet,
  title={Parallelizing Linear Transformers with the Delta Rule over Sequence Length},
  author={Songlin Yang and Bailin Wang and Yu Zhang and Yikang Shen and Yoon Kim},
  journal={ArXiv},
  year={2024},
  volume={abs/2406.06484},
  url={https://api.semanticscholar.org/CorpusID:270371554}
}

@article{yang2025gdn,
  title={Gated Delta Networks: Improving Mamba2 with Delta Rule},
  author={Songlin Yang and Jan Kautz and Ali Hatamizadeh},
  journal={ArXiv},
  year={2024},
  volume={abs/2412.06464},
  url={https://api.semanticscholar.org/CorpusID:274598177}
}

@article{ghafoorian2026rehyat,
  title={ReHyAt: Recurrent Hybrid Attention for Video Diffusion Transformers},
  author={Mohsen Ghafoorian and Amirhossein Habibian},
  journal={ArXiv},
  year={2026},
  volume={abs/2601.04342},
  url={https://api.semanticscholar.org/CorpusID:284543807}
}

@article{fang2025salad,
  title={SALAD: Achieve High-Sparsity Attention via Efficient Linear Attention Tuning for Video Diffusion Transformer},
  author={Tongcheng Fang and Hanling Zhang and Rui Xie and Zhu Han and Xin Tao and Tianchen Zhao and Pengfei Wan and Wenbo Ding and Wanli Ouyang and Xuefei Ning and Yu Wang},
  journal={ArXiv},
  year={2026},
  volume={abs/2601.16515},
  url={https://api.semanticscholar.org/CorpusID:284981782}
}

@article{yu2025videossm,
  title={VideoSSM: Autoregressive Long Video Generation with Hybrid State-Space Memory},
  author={Yifei Yu and Xiaoshan Wu and Xinting Hu and Tianxi Hu and Yang-Tian Sun and Xiaoyang Lyu and Bo Wang and Lin Ma and Yuewen Ma and Zhongrui Wang and Xiaojuan Qi},
  journal={ArXiv},
  year={2025},
  volume={abs/2512.04519},
  url={https://api.semanticscholar.org/CorpusID:283557042}
}

@article{ghafoorian2025surgery,
  title={Attention Surgery: An Efficient Recipe to Linearize Your Video Diffusion Transformer},
  author={Mohsen Ghafoorian and Denis Korzhenkov and Amirhossein Habibian},
  journal={ArXiv},
  year={2025},
  volume={abs/2509.24899},
  url={https://api.semanticscholar.org/CorpusID:281675646}
}

@article{chen2026halo,
  title={Hybrid Linear Attention Done Right: Efficient Distillation and Effective Architectures for Extremely Long Contexts},
  author={Yingfa Chen and Zhen Leng Thai and Zihan Zhou and Zhu Zhang and Xingyu Shen and Shuo Wang and Chaojun Xiao and Xu Han and Zhiyuan Liu},
  journal={ArXiv},
  year={2026},
  volume={abs/2601.22156},
  url={https://api.semanticscholar.org/CorpusID:285140858}
}

@article{li2025klguided,
  title={Distilling to Hybrid Attention Models via KL-Guided Layer Selection},
  author={Yanhong Li and Songlin Yang and Shawn Tan and Mayank Mishra and Rameswar Panda and Jiawei Zhou and Yoon Kim},
  journal={ArXiv},
  year={2025},
  volume={abs/2512.20569},
  url={https://api.semanticscholar.org/CorpusID:284133045}
}

@article{chen2026sanavideo,
  title={SANA-Video: Efficient Video Generation with Block Linear Diffusion Transformer},
  author={Junsong Chen and Yuyang Zhao and Jincheng Yu and Ruihang Chu and Junyu Chen and Shuai Yang and Xianbang Wang and Yicheng Pan and Daquan Zhou and Huan Ling and Haozhe Liu and Hongwei Yi and Hao Zhang and Muyang Li and Yukang Chen and Han Cai and Sanja Fidler and Ping Luo and Song Han and Enze Xie},
  journal={ArXiv},
  year={2025},
  volume={abs/2509.24695},
  url={https://api.semanticscholar.org/CorpusID:281674126}
}

@article{yang2026tempcache,
  title={Fast Autoregressive Video Diffusion and World Models with Temporal Cache Compression and Sparse Attention},
  author={Dvir Samuel and Issar Tzachor and Matan Levy and Micah Green and Gal Chechik and Rami Ben-Ari},
  journal={ArXiv},
  year={2026},
  volume={abs/2602.01801},
  url={https://api.semanticscholar.org/CorpusID:285271188}
}

@article{lv2026lightforcing,
  title={Light Forcing: Accelerating Autoregressive Video Diffusion via Sparse Attention},
  author={Chengtao Lv and Yumeng Shi and Yushi Huang and Ruihao Gong and Shen Ren and Wenya Wang},
  journal={ArXiv},
  year={2026},
  volume={abs/2602.04789},
  url={https://api.semanticscholar.org/CorpusID:285286408}
}

@inproceedings{vaswani2017attention,
  title={Attention is All you Need},
  author={Ashish Vaswani and Noam Shazeer and Niki Parmar and Jakob Uszkoreit and Llion Jones and Aidan N. Gomez and Lukasz Kaiser and Illia Polosukhin},
  booktitle={Neural Information Processing Systems},
  year={2017},
  url={https://api.semanticscholar.org/CorpusID:13756489}
}

@inproceedings{oshima2024ssm,
  title={SSM Meets Video Diffusion Models: Efficient Long-Term Video Generation with Structured State Spaces},
  author={Yuta Oshima and Shohei Taniguchi and Masahiro Suzuki and Yutaka Matsuo},
  year={2024},
  url={https://api.semanticscholar.org/CorpusID:272397789}
}

@article{po2025longctx,
  title={Long-Context State-Space Video World Models},
  author={Ryan Po and Yotam Nitzan and Richard Zhang and Berlin Chen and Tri Dao and Eli Shechtman and Gordon Wetzstein and Xun Huang},
  journal={2025 IEEE/CVF International Conference on Computer Vision (ICCV)},
  year={2025},
  pages={8733-8744},
  url={https://api.semanticscholar.org/CorpusID:278911218}
}

@article{hong2025hth,
  title={Pushing the Boundaries of State Space Models for Image and Video Generation},
  author={Yicong Hong and Long Mai and Yu Yao and Feng Liu},
  journal={ArXiv},
  year={2025},
  volume={abs/2502.00972},
  url={https://api.semanticscholar.org/CorpusID:276094439}
}

@article{zhang2026sla,
  title={SLA: Beyond Sparsity in Diffusion Transformers via Fine-Tunable Sparse-Linear Attention},
  author={Jintao Zhang and Haoxu Wang and Kai Jiang and Shuo Yang and Kaiwen Zheng and Haocheng Xi and Ziteng Wang and Hongzhou Zhu and Min Zhao and Ion Stoica and Joseph E. Gonzalez and Jun Zhu and Jianfei Chen},
  journal={ArXiv},
  year={2025},
  volume={abs/2509.24006},
  url={https://api.semanticscholar.org/CorpusID:281675722}
}

@article{guo2026dummyforcing,
  title={Efficient Autoregressive Video Diffusion with Dummy Head},
  author={Hang Guo and Zhaoyang Jia and Jiahao Li and Bin Li and Yuanhao Cai and Jiangshan Wang and Yawei Li and Yan Lu},
  journal={ArXiv},
  year={2026},
  volume={abs/2601.20499},
  url={https://api.semanticscholar.org/CorpusID:285101812}
}

@article{yin2025causvid,
  title={From Slow Bidirectional to Fast Autoregressive Video Diffusion Models},
  author={Tianwei Yin and Qiang Zhang and Richard Zhang and William T. Freeman and Fr{\'e}do Durand and Eli Shechtman and Xun Huang},
  journal={2025 IEEE/CVF Conference on Computer Vision and Pattern Recognition (CVPR)},
  year={2024},
  pages={22963-22974},
  url={https://api.semanticscholar.org/CorpusID:274610175}
}

@article{sand2025magi,
  title={MAGI-1: Autoregressive Video Generation at Scale},
  author={Sand. ai and Han Sack Teng and Hong Jia and Lei Sun and Lingzhi Li and Maolin Li and Min Tang and Shu Han and Tianning Zhang and W. Q. Zhang and Weifeng Luo and Xiaoyang Kang and Yuchen Sun and Yue Cao and Yunpeng Huang and Yutong Lin and Yuxin Fang and Zewei Tao and Zheng Zhang and Zhongshu Wang and Zixun Liu and Daiqi Shi and Guoli Su and Hanwen Sun and Hong-Fei Pan and Jie Wang and Jie Sheng and Mingyan Cui and Min Hu and Ming Yan and Shucheng Yin and Si-Hui Zhang and Tingting Liu and Xi Yin and Xiaoyu Yang and Xin Song and Xuan Hu and Yankai Zhang and Yu-Qian Li},
  journal={ArXiv},
  year={2025},
  volume={abs/2505.13211},
  url={https://api.semanticscholar.org/CorpusID:278768832}
}

@article{xi2026qvg,
  title={Quant VideoGen: Auto-Regressive Long Video Generation via 2-Bit KV-Cache Quantization},
  author={Haocheng Xi and Shuo Yang and Yilong Zhao and Muyang Li and Han Cai and Xingyang Li and Yujun Lin and Zhuoyang Zhang and Jintao Zhang and Xiuyu Li and Zhiying Xu and Jun Wu and Chenfeng Xu and Ion Stoica and Song Han and Kurt Keutzer},
  journal={ArXiv},
  year={2026},
  volume={abs/2602.02958},
  url={https://api.semanticscholar.org/CorpusID:285275510}
}

@article{yang2025longlive,
  title={LongLive: Real-time Interactive Long Video Generation},
  author={Shuai Yang and Wei Huang and Ruihang Chu and Yicheng Xiao and Yuyang Zhao and Xianbang Wang and Muyang Li and Enze Xie and Ying-Cong Chen and Yao Lu and Song Han and Yukang Chen},
  journal={ArXiv},
  year={2025},
  volume={abs/2509.22622},
  url={https://api.semanticscholar.org/CorpusID:281658815}
}

@article{chen2026pafukv,
  title={Past- and Future-Informed KV Cache Policy with Salience Estimation in Autoregressive Video Diffusion},
  author={Hanmo Chen and Chenghao Xu and Xu Yang and Xuan Chen and Cheng Deng},
  journal={ArXiv},
  year={2026},
  volume={abs/2601.21896},
  url={https://api.semanticscholar.org/CorpusID:285139913}
}

@inproceedings{xu2026kvcachesurvey,
  title={KV Cache Optimization Strategies for Scalable and Efficient LLM Inference},
  author={Yichun Xu and Navjot K. Khaira and Tejinder Singh},
  year={2026},
  url={https://api.semanticscholar.org/CorpusID:286762310}
}

@article{hacohen2024ltx,
  title={LTX-Video: Realtime Video Latent Diffusion},
  author={Yoav HaCohen and Nisan Chiprut and Benny Brazowski and Daniel Shalem and David-Pur Moshe and Eitan Richardson and E. I. Levin and Guy Shiran and Nir Zabari and Ori Gordon and Poriya Panet and Sapir Weissbuch and Victor Kulikov and Yaki Bitterman and Zeev Melumian and Ofir Bibi},
  journal={ArXiv},
  year={2024},
  volume={abs/2501.00103},
  url={https://api.semanticscholar.org/CorpusID:275212083}
}

@article{chen2025skyreels,
  title={SkyReels-V2: Infinite-length Film Generative Model},
  author={Guibin Chen and Dixuan Lin and Jiangping Yang and Chunze Lin and Juncheng Zhu and Mingyuan Fan and Hao Zhang and Sheng Chen and Zhenghao Chen and Chengcheng Ma and Weiming Xiong and Wei Wang and Nuo Pang and Kang Kang and Zhi-Xin Xu and Yuzhe Jin and Yu Liang and Yu-Ning Song and Peng Zhao and Bo Xu and Di Qiu and Debang Li and Zhengcong Fei and Yang Li and Yahui Zhou},
  journal={ArXiv},
  year={2025},
  volume={abs/2504.13074},
  url={https://api.semanticscholar.org/CorpusID:277856899}
}

@article{fan2025rala,
  title={Breaking the Low-Rank Dilemma of Linear Attention},
  author={Qihang Fan and Huaibo Huang and Ran He},
  journal={2025 IEEE/CVF Conference on Computer Vision and Pattern Recognition (CVPR)},
  year={2024},
  pages={25271-25280},
  url={https://api.semanticscholar.org/CorpusID:273969893}
}

@article{li2026packcache,
  title={PackCache: A Training-Free Acceleration Method for Unified Autoregressive Video Generation via Compact KV-Cache},
  author={Kunyang Li and Mubarak Shah and Yuzhang Shang},
  journal={ArXiv},
  year={2026},
  volume={abs/2601.04359},
  url={https://api.semanticscholar.org/CorpusID:284544089}
}

@article{huang2024vbench,
  title={VBench: Comprehensive Benchmark Suite for Video Generative Models},
  author={Ziqi Huang and Yinan He and Jiashuo Yu and Fan Zhang and Chenyang Si and Yuming Jiang and Yuanhan Zhang and Tianxing Wu and Qin Jin and Nattapol Chanpaisit and Yaohui Wang and Xinyuan Chen and Limin Wang and Dahua Lin and Yu Qiao and Ziwei Liu},
  journal={2024 IEEE/CVF Conference on Computer Vision and Pattern Recognition (CVPR)},
  year={2023},
  pages={21807-21818},
  url={https://api.semanticscholar.org/CorpusID:265506207}
}

@article{wang2024vidprom,
  title={VidProM: A Million-scale Real Prompt-Gallery Dataset for Text-to-Video Diffusion Models},
  author={Wenhao Wang and Yi Yang},
  journal={ArXiv},
  year={2024},
  volume={abs/2403.06098},
  url={https://api.semanticscholar.org/CorpusID:269293420}
}


\newpage
\appendix

\section{Inference algorithm}
\label{app:algorithm}

Algorithm~\ref{alg:hybrid} describes our hybrid inference with a fixed-size recurrent state.

\begin{algorithm}[h]
\caption{ARL$^2$ hybrid inference (ours)}
\label{alg:hybrid}
\begin{algorithmic}[1]
\Require text prompt $p$, number of blocks $N$, denoising steps $T = [t_1, \dots, t_m]$
\Ensure video frames $Y = [y_0, \dots, y_{N-1}]$
\State $S \gets \mathbf{0} \in \mathbb{R}^{H \times D \times D}$ \Comment{fixed-size GDN state per hybrid layer}
\For{$i = 0$ to $N-1$}
    \State $z_i \sim \mathcal{N}(0, I)$
    \For{$t \in T$} \Comment{denoising loop --- state is \textbf{not} updated}
        \State $Q, K, V \gets W_q(z_i),\, W_k(z_i),\, W_v(z_i)$
        \State \textcolor{intraBlue}{$O_{\mathrm{intra}} \gets \mathrm{Softmax}(\mathrm{RoPE}(Q),\, \mathrm{RoPE}(K),\, V)$} \Comment{intra-block only: $L \times L$}
        \State \textcolor{interGreen}{$Q' \gets \mathrm{L2Norm}(\mathrm{RoPE}(\phi_q(Q)))$}
        \State \textcolor{interGreen}{$O_{\mathrm{inter}} \gets Q' \cdot S$} \Comment{query fixed state: $O(H D^2)$}
        \State $O \gets O_{\mathrm{intra}} + G \odot O_{\mathrm{inter}}$ \Comment{gated combination}
        \State $z_i \gets \mathrm{denoise\_step}(z_i, O, t)$
    \EndFor
    \State $\hat{z}_i \gets$ clean latent prediction \Comment{clean pass --- state is updated}
    \State $K, V \gets W_k(\hat{z}_i),\, W_v(\hat{z}_i)$
    \State \textcolor{interGreen}{$K' \gets \mathrm{L2Norm}(\mathrm{RoPE}(\phi_k(K))),\quad V' \gets \phi_v(V)$}
    \State \textcolor{interGreen}{$\alpha \gets \mathrm{gate\_proj}(\hat{z}_i),\quad \beta \gets \mathrm{lr\_proj}(\hat{z}_i)$} \Comment{per-token forget/learning gates}
    \State \textcolor{interGreen}{$S \gets \mathrm{GDN\_update}(S, K', V', \alpha, \beta)$} \Comment{$S$ remains $H {\times} D {\times} D$}
    \State $y_i \gets \mathrm{decode}(\hat{z}_i)$
\EndFor
\State \Return $Y$
\end{algorithmic}
\end{algorithm}

Algorithm~\ref{alg:hybrid} highlights the two key design principles of ARL$^2$: (1)~\textbf{Frame-level access}: during each denoising step, all $L$ tokens in the current block query the same pre-update state $S$, ensuring consistent history observation (lines 7--8); (2)~\textbf{Clean-pass update}: the GDN state is updated only after the final clean pass (line 13), preventing noisy denoising intermediates from contaminating the persistent memory. The state $S$ has a fixed size of $H \times D \times D$ regardless of the number of generated blocks, reducing per-layer memory from $O(N)$ to $O(1)$.

\section{Evaluation details}
\label{app:evaluation}


\paragraph{Efficiency evaluation.}
Efficiency measurements (Table~2 in the main text) are conducted on a single NVIDIA RTX PRO 6000 Blackwell (96\,GB). Reported generation time (\texttt{gen\_s}) measures end-to-end video generation (denoising loop + VAE decode), excluding text encoding and model loading. Memory is \texttt{torch.cuda.max\_memory\_allocated()}, reset per video. All runs use cold-start conditions (GPU $\le 50\,^{\circ}$C) to eliminate thermal throttling bias. Self Forcing is architecturally identical to Causal Forcing (only EMA weights differ), so their latency and memory are equal.

\paragraph{Stage~1 stress-test evaluation.}
Layer selection uses a curated set of stress-test prompts (Table~\ref{tab:stress-prompts}) designed to maximally expose recurrent state weaknesses across three capability axes: temporal consistency (TC), motion dynamics (DY), and spatial quality (SQ). Each of the 30 layers is evaluated via single-layer replacement on 6 VBench quality dimensions. 
The resulting per-layer sensitivity scores guide which layers are safe for hybrid replacement.

\begin{table}[h]
\centering
\small
\setlength{\tabcolsep}{4pt}
\caption{Stress-test prompts for sensitivity-guided layer selection. Each prompt targets a specific state weakness axis: TC = temporal consistency, DY = motion dynamics, SQ = spatial quality.}
\label{tab:stress-prompts}
\resizebox{\textwidth}{!}{%
\begin{tabular}{clll}
\toprule
ID & Target Axis & Stress Category & Prompt (abbreviated) \\
\midrule
S00 & TC & Appearance retention & Woman with red striped scarf walks down cobblestone street \\
S01 & TC & Brief occlusion & Golden retriever runs behind oak tree and re-emerges \\
S02 & TC & Long occlusion & White SUV passes behind rocky outcropping and re-emerges \\
S03 & TC & Background stability & Ballet dancer performs pirouette on ornate theater stage \\
S04 & TC & Multi-entity tracking & Two puppies (golden retriever + black labrador) play on beach \\
S05 & DY & Fast motion & Hummingbird darts between tropical flowers \\
S06 & TC+DY & Camera orbit & Camera slowly orbits white marble statue in museum gallery \\
S07 & SQ & Static scene & Stop motion of flower growing on windowsill \\
S08 & SQ & Slow portrait & Victoria crowned pigeon close-up with blue plumage \\
S09 & DY+TC & Natural dynamics & Train traveling through Tokyo suburbs with reflections \\
S10 & DY & Group motion & Flock of paper airplanes through dense jungle canopy \\
S11 & TC+SQ & Reflection + motion & Gray-haired man at Parisian caf\'{e}, cinematic lighting \\
\bottomrule
\end{tabular}%
}
\end{table}


\section{Qualitative comparison across models}
\label{app:qualitative}

\begin{figure}[h]
  \centering
  \includegraphics[width=\textwidth]{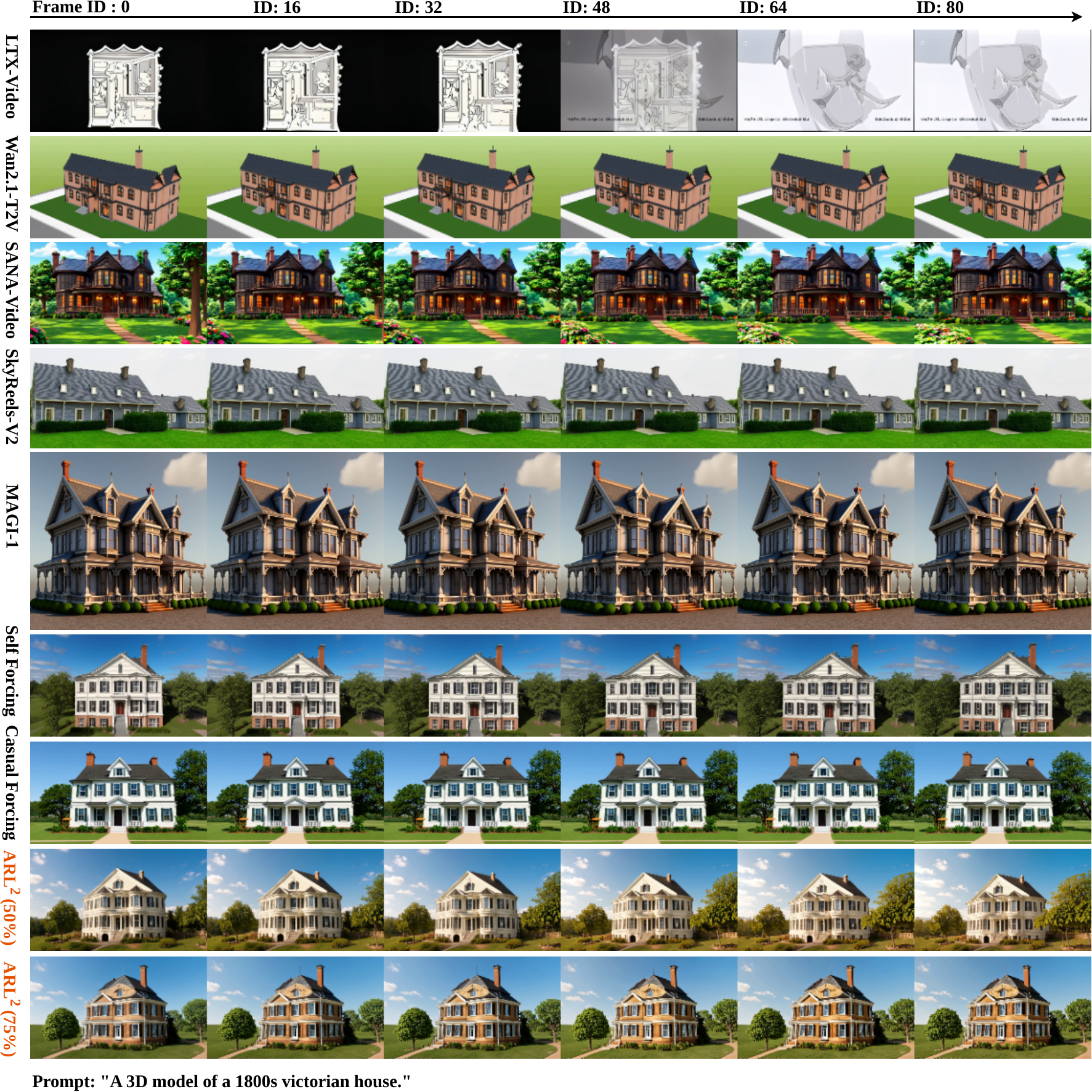}
  \caption{Qualitative comparison across all evaluated models. Six uniformly sampled frames from 81-frame videos generated with the prompt \emph{``A 3D model of a 1800s victorian house.''} ARL$^2$ variants maintain consistent appearance and structural detail comparable to the full-softmax baselines while operating with significantly reduced memory and compute.}
  \label{fig:vis-supply}
\end{figure}

Figure~\ref{fig:vis-supply} provides a side-by-side qualitative comparison of all evaluated models on the prompt \emph{``A 3D model of a 1800s victorian house.''} We uniformly sample 6 frames (IDs 0, 16, 32, 48, 64, 80) from each 81-frame generation. LTX-Video loses structural coherence after the first few frames, collapsing into unrecognizable artifacts. Wan2.1-T2V and SANA-Video produce stylistically consistent results but exhibit limited 3D structure. SkyReels-V2 maintains temporal stability but generates a flat, low-detail scene. MAGI-1 renders a detailed Victorian house with strong temporal consistency, benefiting from its larger 4.5B parameter count. Among the 1.3B-class autoregressive models, Self Forcing and Causal Forcing both produce plausible houses. ARL$^2$ (50\%) achieves visual quality comparable to them, with stable appearance and consistent architectural details throughout the sequence, while running $1.40\times$ faster and using $31\%$ less memory. ARL$^2$ (75\%) maintains coherent structure and lighting across all frames with an even larger efficiency margin: $2.26\times$ speedup and $54\%$ memory reduction, laying a strong foundation for streaming video generation. Playable video results are provided in the supplementary zip file.

\section{Detailed VBench per-dimension results}
\label{app:full-vbench}

Table~\ref{tab:vbench-full-946} reports the detailed VBench scores evaluated on the full 946-prompt official benchmark. All models generate at $832\times480$, 81 frames unless noted.

\begin{table}[h]
\centering
\small
\setlength{\tabcolsep}{3pt}
\caption{Detailed VBench per-dimension scores. $\ddagger$: native resolution 544$\times$960, 97 frames.}
\label{tab:vbench-full-946}
\resizebox{\textwidth}{!}{%
\begin{tabular}{l ccccccccc}
\toprule
Dimension & \shortstack{SANA-\\Video} & \shortstack{Wan2.1\\T2V} & \shortstack{LTX-\\Video} & CF & SF & \shortstack{ARL$^2$\\(50\%)} & \shortstack{ARL$^2$\\(75\%)} & \shortstack{MAGI-1$^\dagger$} & \shortstack{SkyReels$^\ddagger$} \\
\midrule
\multicolumn{10}{l}{\emph{Quality dimensions}} \\
subject\_consistency & 97.09 & 96.18 & 87.89 & 96.19 & 95.62 & 95.63 & 94.85 & 95.71 & \textbf{97.78} \\
background\_consistency & 96.95 & \textbf{98.33} & 92.00 & 96.44 & 96.40 & 96.41 & 95.94 & 97.14 & 97.68 \\
temporal\_flickering & 89.22 & \textbf{98.33} & 98.20 & 96.27 & 96.82 & 97.26 & 96.90 & 97.69 & 97.97 \\
motion\_smoothness & 95.10 & 96.60 & 97.91 & 94.24 & 95.59 & 95.59 & 94.24 & \textbf{98.67} & 98.50 \\
aesthetic\_quality & \textbf{70.48} & 65.35 & 48.08 & 68.80 & 66.75 & 68.29 & 68.44 & 62.44 & 59.66 \\
imaging\_quality & 67.55 & 68.64 & 58.62 & \textbf{70.62} & 69.96 & 69.87 & 69.87 & 64.84 & 62.29 \\
\midrule
\multicolumn{10}{l}{\emph{Semantic dimensions}} \\
object\_class & \textbf{94.86} & 75.00 & 29.30 & 85.13 & 84.89 & 86.23 & 83.62 & 85.76 & 32.03 \\
multiple\_objects & \textbf{81.78} & 64.12 & 8.20 & 64.94 & 68.29 & 67.00 & 68.29 & 50.00 & 15.63 \\
human\_action & 92.00 & 62.50 & 62.50 & 79.00 & 80.00 & 77.00 & 72.00 & \textbf{94.44} & 50.00 \\
color & 88.12 & 85.71 & 97.62 & 90.72 & 93.98 & 91.48 & 91.75 & 86.46 & \textbf{100.00} \\
spatial\_relationship & 77.46 & 57.92 & 67.88 & \textbf{83.43} & 78.45 & 78.74 & 78.62 & 76.24 & 38.77 \\
scene & \textbf{73.36} & 26.48 & 15.69 & 34.65 & 32.79 & 32.97 & 36.15 & 27.03 & 15.69 \\
appearance\_style & 76.86 & 66.98 & 73.71 & 66.81 & 66.77 & 65.66 & 67.76 & \textbf{87.12} & 66.47 \\
temporal\_style & \textbf{67.72} & 63.40 & 44.02 & 66.10 & 64.43 & 65.65 & 66.83 & 64.38 & 55.58 \\
overall\_consistency & 74.70 & 64.99 & 62.97 & 67.98 & 66.80 & 66.43 & 67.16 & \textbf{76.74} & 58.00 \\
\midrule
\textbf{Quality Avg} & 86.07 & \textbf{87.24} & 80.45 & 87.09 & 86.86 & 87.17 & 86.71 & 86.08 & 85.65 \\
\textbf{Semantic Avg} & \textbf{80.76} & 63.01 & 51.32 & 70.97 & 70.71 & 70.13 & 70.24 & 72.02 & 48.02 \\
\textbf{Total} & \textbf{85.00} & 82.39 & 74.62 & 83.87 & 83.63 & 83.77 & 83.41 & 83.27 & 78.12 \\
\bottomrule
\end{tabular}%
}
\end{table}


\section{Broader impact}
\label{app:impact-limitations}

ARL$^2$ reduces the computational and memory cost of autoregressive video generation by up to $2.26\times$ and $54\%$, respectively. This efficiency gain lowers the hardware barrier for long-form video synthesis, potentially democratizing access to high-quality video generation for researchers and practitioners with limited GPU resources. By enabling longer videos on a single consumer-grade GPU, the method may accelerate applications in education, content creation, and accessibility tools. However, as with all generative video models, the technology could be misused to produce misleading or harmful synthetic media. We encourage the community to develop and deploy appropriate watermarking, provenance tracking, and content authentication mechanisms alongside efficient generation methods.



\newpage

\end{document}